\documentclass{article}

\usepackage{microtype}
\usepackage{graphicx}
\usepackage{subfigure}
\usepackage{booktabs}
\usepackage{hyperref}

\usepackage[accepted]{icml2025}
\usepackage{amsmath}
\usepackage{amssymb}
\usepackage{mathtools}
\usepackage{amsthm}
\usepackage[capitalize,noabbrev]{cleveref}
\usepackage[textsize=tiny]{todonotes}

\icmltitlerunning{SAGDA: Open-Source Synthetic Agriculture Data for Africa}

\begin{document}

\twocolumn[
\icmltitle{SAGDA: Open-Source Synthetic Agriculture Data for Africa}

\begin{icmlauthorlist}
\icmlauthor{Abdelghani Belgaid}{um6p}
\icmlauthor{Oumnia Ennaji}{um6p}
\end{icmlauthorlist}

\icmlaffiliation{um6p}{College of Computing, Mohammed VI Polytechnic University, Lot 660, Ben Guerir, 43150, Morocco}

\icmlcorrespondingauthor{Abdelghani Belgaid}{Abdelghani.BELGAID@um6p.ma}
\icmlcorrespondingauthor{Oumnia Ennaji}{Oumnia.ENNAJI@um6p.ma}

\icmlkeywords{Open-Source, Synthetic Data, Machine Learning, Agriculture, Africa}

\vskip 0.3in
]

\printAffiliationsAndNotice{\icmlEqualContribution}

\begin{abstract}
Data scarcity in African agriculture hampers machine learning (ML) model performance, limiting innovations in precision agriculture. The Synthetic Agriculture Data for Africa (SAGDA) library, a Python-based open-source toolkit, addresses this gap by generating, augmenting, and validating synthetic agricultural datasets. We present SAGDA’s design and development practices, highlighting its core functions: generate, model, augment, validate, visualize, optimize, and simulate, as well as their roles in applications of ML for agriculture. Two use cases are detailed: yield prediction enhanced via data augmentation, and multi-objective NPK (nitrogen, phosphorus, potassium) fertilizer recommendation. We conclude with future plans for expanding SAGDA’s capabilities, underscoring the vital role of open-source, data-driven practices for African agriculture.
\end{abstract}

\section{Introduction}
African agriculture faces a data scarcity challenge that impedes ML-driven insights\cite{Paudel2021,ennaji2023machine,Ennaji2025}. While Africa holds unique agronomic data, access is often restricted by fragmentation, privacy, and limited collection efforts. This dearth of local data stifles AI’s potential in improving crop yields, as ML models trained on insufficient or skewed datasets may be biased or inaccurate. Synthetic data generation has emerged as a promising solution: by simulating statistically realistic agricultural data, researchers and practitioners can augment limited datasets without compromising privacy \cite{akkem2024comprehensive,chia2022relationprompt,xu2018synthesizing,xu2019modeling}. \cite{gartner2022syntheticdata} predicts that synthetic data will outpace real data in AI by 2030, highlighting a global shift toward generated datasets. In Africa, synthetic data offers a path to 'democratize' AI ensuring that the benefits of ML reach communities lacking robust datasets.

SAGDA (Synthetic Agriculture Data for Africa) \cite{SAGDA} \footnote{Code available at \url{https://github.com/SAGDAfrica/sagda}} is an open-source initiative responding to this need. It provides tools to generate synthetic climate records, soil profiles, crop yield series, and fertilizer usage data, among others, all tailored to African agricultural contexts to enable a realistic simulation of farming conditions. The motivation is twofold: to unlock agricultural ML innovation by bridging data gaps, and to ensure this solution remains accessible and open-source, fostering community-driven improvements.

\section{System Overview of SAGDA}
SAGDA is implemented as a Python library (available on GitHub and PyPI) built with modular components for synthetic data workflows. The list below gives an overview of its core modules and functions, illustrating how each addresses aspects of agricultural ML pipelines.

  \textbf{Dataset (sagda.data):} Provides streamlined access to real and synthetic datasets used in agricultural ML workflows. Includes preprocessed public sources and selected crowdsourced datasets from African institutions. Enables users to query datasets by type, region, or time range, and to reuse outputs from synthetic generation. This module supports reproducibility, lowers entry barriers, and ensures consistency across tasks.

  \textbf{Data Generation (sagda.generate):} Creates synthetic agricultural datasets from scratch. For example, it can synthesize time-series weather data for a region or simulate a distribution of soil nutrients and crop yields given parameter ranges. Under the hood, this module leverages statistical distributions and generative models (e.g., variational autoencoders, GANs) to produce data that mimics real-world patterns. SAGDA’s generation function supports geospatial and temporal variations, enabling outputs that reflect seasonal rainfall cycles or spatial soil fertility gradients.

  \textbf{Model (sagda.model):} Provides access to pretrained models for key agricultural tasks such as yield prediction and fertilizer optimization. These models are trained on curated datasets and include built-in preprocessing and evaluation metrics. Users can apply them directly to real or synthetic data for benchmarking, validation, or scenario analysis. This module helps assess the practical utility of generated data, supporting reproducible research and accelerating adoption in applied settings.
  
  \textbf{Data Augmentation (sagda.augment):} Enhances existing datasets by creating new synthetic samples. SAGDA can apply methods such as random sampling, linear regression extrapolation, and even neural network-based generation (autoencoders or GANs) to augment small datasets. For instance, if a dataset has few samples of drought years, SAGDA can augment it with synthetic drought-year data points to balance the dataset. Augmentation ensures ML models get diverse and balanced training data, mitigating bias from class imbalance. In an agricultural context, this means models can better handle rare but critical scenarios (e.g., extreme droughts or pest outbreaks).
  
  \textbf{Data Validation (sagda.validate):} A crucial step where synthetic data is compared against real data to ensure realism and utility. SAGDA’s validation suite computes statistical similarity metrics (e.g., Kolmogorov–Smirnov tests, distribution overlap measures) to quantify how closely synthetic data resembles real datasets. It also includes domain-specific checks—like ensuring generated weather sequences respect known climate patterns or that synthetic soil profiles are physically plausible (e.g., pH in a valid range). This module helps users iteratively refine their synthetic data generation process, flagging anomalies or biases introduced during generation.
  
  \textbf{Data Visualization (sagda.visualize):} To interpret and communicate synthetic data, SAGDA provides plotting functions (e.g., distribution plots, time-series trends, spatial heatmaps). Visualization helps users qualitatively assess synthetic data realism and compare it side-by-side with real data. For example, one can visualize crop yield distributions from synthetic farms versus actual survey data to check alignment in range and variance.

  \textbf{Optimization (sagda.optimize):} This module supports creating and evaluating synthetic scenarios for agricultural optimization problems. For instance, SAGDA can simulate various fertilizer application strategies across a synthetic farm dataset and evaluate yield outcomes. By integrating with optimization algorithms (e.g., simulated annealing or particle swarm optimization), SAGDA helps identify optimal decisions such as ideal NPK fertilizer combinations under certain synthetic conditions (e.g., nutrient use, economic viability, environmental impact).

  \textbf{Simulation (sagda.simulate):} Beyond static data generation, SAGDA’s simulation capabilities allow dynamic modeling of agricultural processes. For example, users can simulate a growing season day by day, with interactions between weather events, crop growth stages, and management actions. This function can ingest synthetic or real initial conditions and simulate forward, producing sequential data useful for time-series ML models or scenario analysis. Integration with external APIs (e.g., NASA POWER, OpenWeatherMap) allows real data to be blended with synthetic simulations, grounding simulations in reality. With SAGDA, users can generate synthetic fields and test what-if scenarios in a risk-free environment, accelerating agronomic insights without requiring immediate field trials.

SAGDA’s architecture prioritizes modularity and extensibility. Each core function (generate, augment, etc.) is encapsulated in its module (e.g., data\_generation.py, data\_augmentation.py) with clear APIs. This separation allows contributors to add new algorithms or data types (e.g., adding an algorithm in the generation module) without breaking others. The library has minimal external dependencies (pandas, numpy, scipy, tensorflow, scikit-learn) to ease installation and foster integration into existing ML pipelines. As a PyPI package (pip install sagda), it’s accessible to a broad user base.

\section{Use Cases}
We demonstrate SAGDA’s capabilities through two representative use cases: crop yield prediction with data augmentation and NPK fertilizer optimization. These examples, drawn from real scenarios, show how SAGDA’s synthetic data functions integrate into ML workflows and contribute to improved outcomes.

\subsection{Crop Yield Prediction with Augment Method}
\textbf{Scenario:} Accurate crop yield prediction is vital for food security planning. However, yield datasets in many African settings suffer from temporal imbalance—certain seasons have sparse data due to limited monitoring or inconsistent field trials. In our case, over 70\% of the data came from the most recent season (2020–2021), while earlier seasons (2018–2020) were underrepresented. This imbalance limited model generalization to temporally distant conditions. To address this, SAGDA’s augment() method was applied to oversample training data from the earlier seasons which internally applies SMOTE \cite{chawla2002smote}, increasing representation within low-yield quartiles and enhancing the temporal coverage for modeling.

\textbf{Model Performance:} The augmented dataset is then used to train multiple machine learning models under a temporal holdout setup, simulating real-world deployment on future seasons. By comparing performance with and without augmentation, we evaluate SAGDA’s impact. The augmentation increased training data volume by 145.2\%. Among all configurations, a stacked ensemble with a linear meta-learner achieved the best performance, reaching a MAPE of 29.85\% on temporally disjoint test data, significantly outperforming the baseline model trained without augmentation, which reached nearly 40\% MAPE. This result highlights SAGDA’s capacity to mitigate temporal data sparsity and improve yield prediction accuracy under real-world constraints.

\textbf{Synthetic Data Validation:}We validated the quality of the augmented dataset using comparative statistical analysis. Principal Component Analysis (PCA) showed that the first two components explained 13.0\% and 6.6\% of the variance, respectively. A Mahalanobis distance-based comparison confirmed that synthetic and original samples shared a 99.85\% overlap in the PCA space, indicating high structural similarity. These findings suggest that the synthetic data preserved key agronomic patterns while expanding representation across early growing seasons.

\textbf{Discussion:} The yield prediction use case demonstrates how targeted temporal augmentation via SAGDA can enhance model robustness in low-data regimes. By embedding augmentation directly in the library, users can balance temporally skewed datasets without requiring external preprocessing.The improved performance under a temporal split highlights the potential of such methods for yield forecasting across variable agricultural seasons.

\subsection{Site-Specific NPK Fertilizer Optimization}
\textbf{Scenario:} Optimizing fertilizer application is a pressing challenge in precision agriculture. Over- or under-application of nutrients can reduce yields and harm the environment. Data-driven approaches can tailor fertilizer recommendations to local conditions, but require detailed field data. In this use case, we leverage a dataset composed of thousands of field trial observations with soil properties (e.g., pH, organic matter, NPK levels) and yield outcomes. We use SAGDA’s optimize() to explore improved fertilizer strategies.

\textbf{Optimization:} Using the optimize() module, we implemented a multi-objective function that balances yield improvement, nutrient efficiency, and environmental impact. This is done through scenario-based simulation using top-performing predictive models and metaheuristic optimization techniques, such as Simulated Annealing and Particle Swarm Optimization. The module searches for the optimal nitrogen (N), phosphorus (P), and potassium (K) rates tailored to each field, under constraints like total input limits and soil sensitivity.

\textbf{Results:} SAGDA’s optimization module returned consistent improvements across all tested crops and scenarios. On average, yield increased by over 500 kg/ha while maintaining high nutrient use efficiency and moderate environmental impact scores. To ensure agronomic realism, we evaluated the similarity between recommended and observed nutrient levels using an Explained Variability metric, which reached up to 86\% in soft wheat and averaged above 60\% across crops. These findings confirm that SAGDA’s optimized recommendations are both statistically sound and practically aligned with real-world nutrient needs.

These use cases demonstrate how open-source tools like SAGDA complement academic research and practical applications in precision agriculture. In the yield prediction scenario, SAGDA’s augment() method addressed temporal imbalances by generating realistic synthetic samples for underrepresented seasons, significantly improving model accuracy under temporal splits by nearly 10\%. In the site-specific NPK fertilizer optimization, SAGDA’s optimize() module identified yield-maximizing nutrient strategies across diverse soil conditions, improving yields by over 500 kg/ha on average while maintaining nutrient efficiency.

By releasing the synthetic data generation and optimization modules, SAGDA exemplifies OSS principles: reusability, transparency, and scalable adaptation. Researchers and agronomists across Africa can leverage SAGDA to simulate local scenarios and optimize fertilizer recommendations before costly field trials, accelerating innovation in data-scarce regions.

\section{Future Work}
Looking ahead, several avenues will guide the evolution of SAGDA:

  \textbf{Expanded Datasets and Variables:} We aim to incorporate more crops (e.g., tubers, fruit trees) and variables (e.g., pest incidence, market prices) into SAGDA’s generation capabilities. This requires gathering reference data for these new variables to calibrate the synthetic generation. Collaborations with agricultural institutes across Africa can provide domain knowledge for these expansions. Additionally, we plan to integrate remote sensing data as either inputs or outputs for synthetic data, possibly linking with image generation models.

  \textbf{Advanced Simulation Models:} Current simulations are relatively simple or rely on ML black-boxes. We plan to integrate process-based crop models (e.g., DSSAT, APSIM) to simulate agronomic processes more realistically, and to develop ML emulators for faster approximations. We will also explore adversarial training to further enhance data realism, while ensuring that outputs remain consistent with agronomic principles through hybrid modeling approaches (combining mechanistic and ML components). Additionally, we aim to release pre-trained models for agricultural tasks (e.g., yield prediction, fertilizer recommendation, soil property estimation), trained on real and SAGDA-augmented datasets to provide baselines and lower entry barriers.

  \textbf{User Interface \& Accessibility:} To broaden adoption, we plan to develop a simple GUI or web interface for SAGDA. This could be a lightweight dashboard where users select a crop and region, then click to generate or augment data without writing code. Such an interface would make SAGDA useful to agricultural extension officers and students. This aligns with feedback from stakeholders who prefer visual tools over code.
  
  \textbf{Empirical Studies on Impact:} To convince more stakeholders of synthetic data’s value, we plan joint studies with universities: e.g., comparing model performance on real vs. augmented vs. fully synthetic data across multiple African countries. Publishing these results (even negative findings, if any) will guide best practices. For instance, one open question is how much synthetic data is too much in a training set; we will experiment to find optimal mixes.

  \textbf{Integration with Policy and Education:} Beyond research, we want SAGDA to inform policy. Future work involves creating scenario simulators that can help policymakers visualize outcomes (like projecting national yield under climate change using synthetic data when real projections are sparse). On the education side, we plan to include SAGDA in data science curricula focused on agriculture, giving students a hands-on tool to practice generating and using synthetic data.

The future of SAGDA is geared towards being more user-friendly, comprehensive in scope, and deeply validated. Its open-source essence remains central: we will continue to grow its community and encourage spin-offs or extensions that cater to specific needs. We hope SAGDA can serve as a template for similar initiatives in other data-scarce domains.

\section{Conclusion}
We presented SAGDA (Synthetic Agriculture Data for Africa), an open-source library addressing data scarcity in African agriculture through synthetic data generation and augmentation. The library’s core modules—generate, augment, validate, visualize, optimize, simulate—form a pipeline that enriches agricultural ML workflows with realistic synthetic data, as evidenced by our yield prediction and fertilizer optimization use cases. SAGDA’s development underscores the importance of open-source best practices: rigorous testing, documentation, community engagement, and modular design.

Open challenges remain, including ensuring the quality and fairness of synthetic data and fostering broad adoption among non-technical stakeholders. However, the progress so far demonstrates that an open-source tool like SAGDA can meaningfully contribute to data-driven innovation in agriculture. By enabling researchers and practitioners to simulate “what-if” scenarios and augment scarce datasets, SAGDA helps unlock insights that would be impractical or impossible with limited real data alone. This is particularly significant for Africa, where filling data gaps can directly translate to better-informed decisions for food security and sustainable farming.

\section*{Impact Statement}
This work aims to democratize access to agricultural ML tools by enabling realistic synthetic data generation across data-poor African contexts. It contributes to reproducible, inclusive AI research and has the potential to improve food security decisions through better data availability.

\bibliographystyle{icml2025}
\bibliography{references}

\end{document}